# Artificial Brain Based on Credible Neural Circuits in a Human Brain

By John Robert Burger
Professor Emeritus
Department of Electrical and Computer Engineering
25686 Dahlin Road
Veneta, OR 97487
(jrburger1@gmail.com)

***Summary*** – Neurons are individually translated into simple gates to plan a brain based on human psychology and intelligence. State machines, assumed previously learned in subconscious associative memory are shown to enable equation solving and rudimentary thinking using nanoprocessing within short term memory.

## Introduction

The goal of this paper is to plan an artificial brain under the constraints imposed by circuit and system theory, that is, without metaphysical or divine intervention. The plan exploits homomorphism between pulsating neurons and dc logic. The author was inspired to this effort over 55 years ago by a paperback aimed at capturing the imagination of anyone curious and open minded enough to read it (Berkeley, 1949). In Berkeley's words, "What men know about the way in which a human brain thinks can be put down in a few pages, and what men do not know would fill many libraries." With what little we know biologically and psychologically, this paper sets out to synthesize how a human brain likely works.

Visualizing the brain as just another electrical system is in itself somewhat novel given the many impractical brain and long term potentiation schemes that find their way into scientific thought. Clearly specialists in life sciences are not very knowledgeable about the constraints imposed by circuits and systems nor are they are particularly competent at logic synthesis. Acknowledging the role of electrical circuits imposes severe constraints on a brain, especially on structure and performance.

The work below, although a humble first step, at least differs greatly from work with *artificial neural networks* (Haikonen 2003; de Garis, 2010). Other current efforts mentioned in popular science magazines although astonishingly expensive are very far indeed from a practical result (Agee, 2008; IBM, 2010).

A major constraint that must never be forgotten is that the brain is an associative processor that may be modeled with a particular logic structure (Burger 2009). Anything less than associative processing is wrong. If ultimately an artificial brain is to approach human abilities, it has got to be more than a John von Neumann machine. In particular, anthropomorphic computing needs to be organized between short term memory (consciousness) and associative long term memory (subconscious). The psychological characteristics of explicit human memory, conscious and subconscious, are quite well understood and available elsewhere. As a convenience for those who have never taken an interest in the biological and psychological sciences, including my fellow engineering stereotypes, salient facts are mentioned below.

Short and long term memories are built upon attributes such as shape, color, texture, edge locations and many other attributes. Psychologists have yet to compile a complete list of attributes. Still it is possible to understand how human memory works. An attribute may be assumed to be a Boolean signal that takes meaning from its specific fixed location in a row of distributed attributes in short term memory. Short term memory neurons can be modeled as triggered devices that remains activated for up to several seconds. They are energy efficient compared to DRAM, in which mere capacitors discharge dissipatedly through channel resistance.

Not everything about the brain is well understood, but it appears that the activation of memory neurons or any other neurons for signaling involves the application of excitory neurotransmitters within close-spaced synapses, about 20 nm. The output of an activated neuron is a low energy, low frequency <100 Hz burst of low voltage pulses, charging and discharging membrane capacitance between about -70 mV to +40 mV. Bursts from short term memory neurons persist for seconds, while ordinary logic neurons for signaling need only a few tens of milliseconds for a burst of about ten pulses propagating down their axons. Neural logic involves pre and post synaptic triggering with excitory neurotransmitters, dendritic processing, inhibitory neurotransmitters, specialized geometries, thresholds, soma activation and axon pulse bursts, all of which are referred to glibly, if not superficially in the literature as an *action potential*.

It is generally acknowledged by even the most intractable reviewers that neurons are capable of arbitrary Boolean logic (Burger Sept 23, 2010). To achieve a Boolean definition, inputs and outputs are defined to be local true or false releases of neurotransmitters. Although there may be many specialized neurons, only three types are necessary for an artificial brain, as illustrated in Fig. 1. Ordinary logic neurons, whose outputs persist tens of milliseconds, short term memory neurons whose outputs persist up to many seconds, and long term memory neurons, whose outputs persist indefinitely.

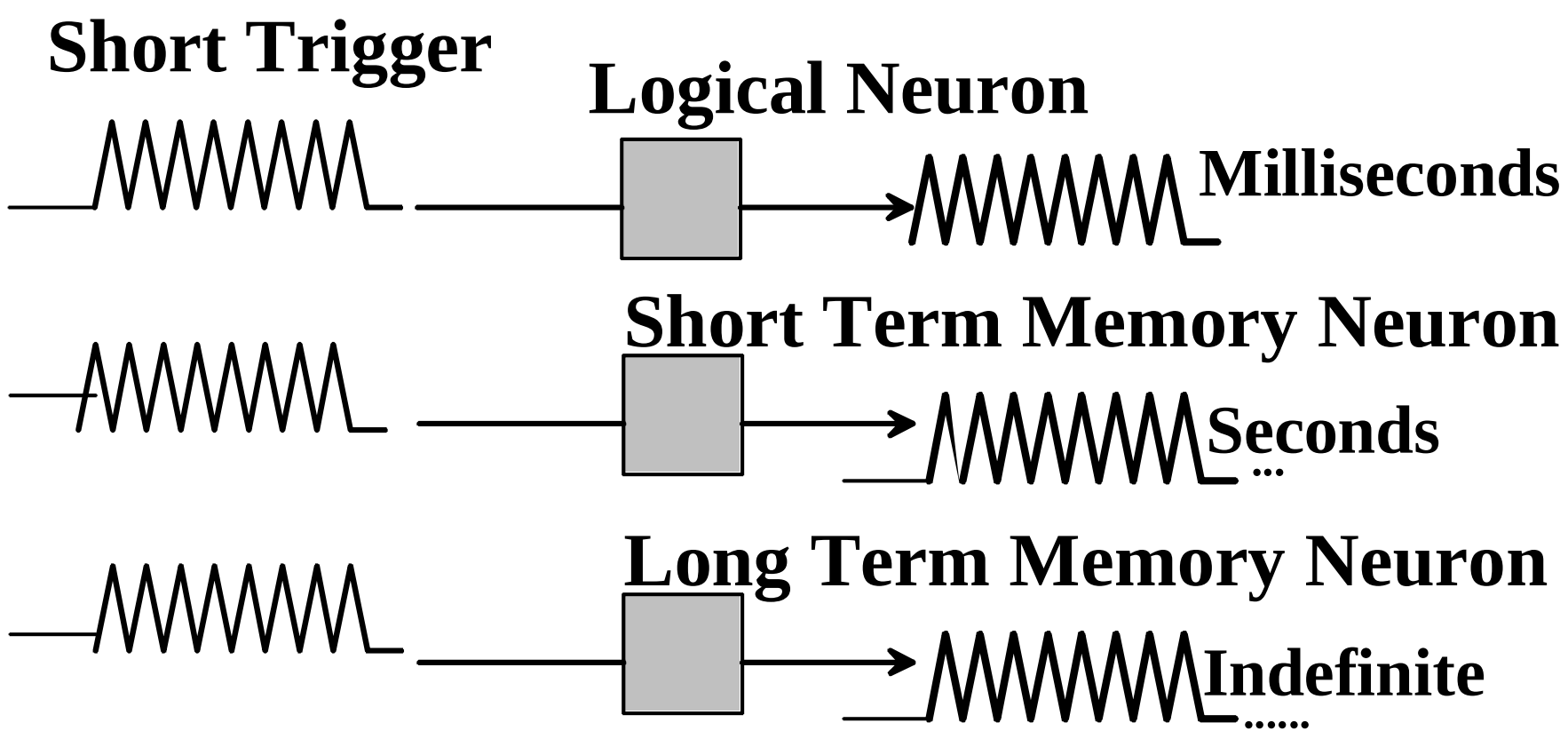

**Fig. 1 Neurons needed for signaling**

A long term memory neuron can be viewed as an efficient programmable ROM. As in human memory, a PROM, once programmed, cannot be erased or overwritten. A biological PROM with pulsating input for triggering and pulsating output to signal a memory attribute, is depicted in Fig. 2a. The output is fed back to an input so that once the pulses are triggered, they cycle indefinitely. It is important to note that a neural ring oscillator can, and must be energy efficient.

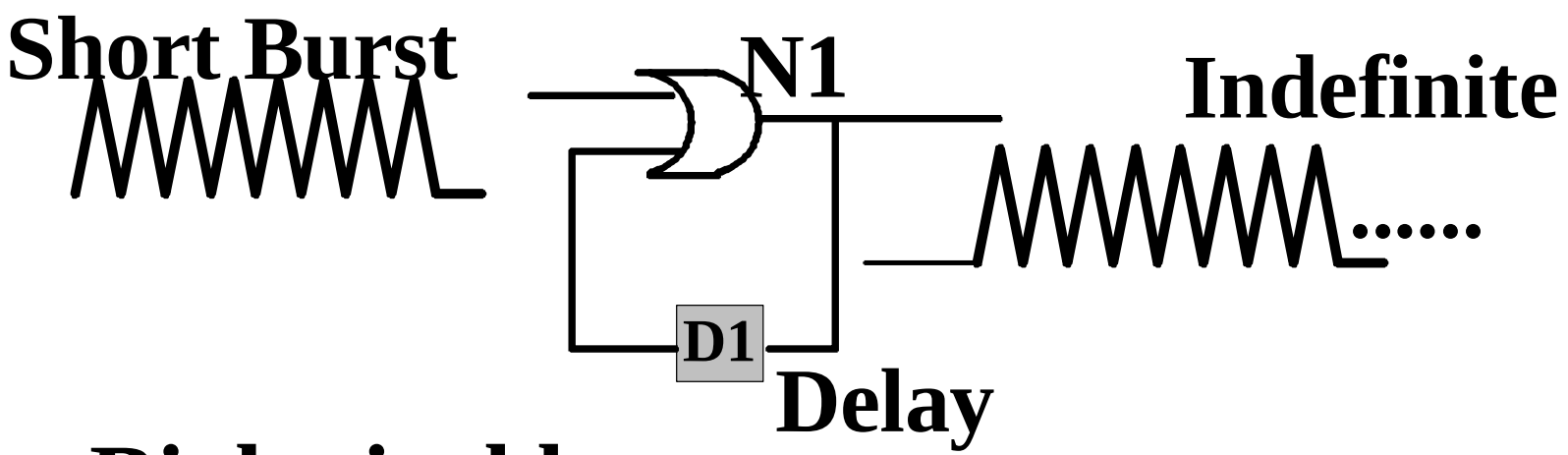

**(a) Biological long term memory**

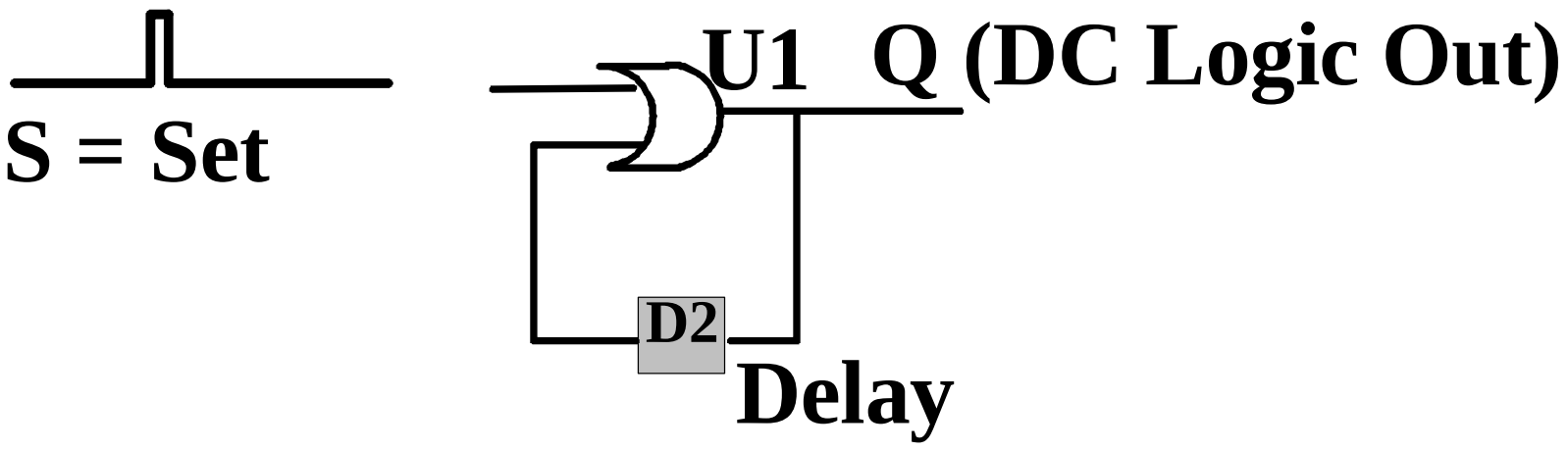

**(b) CMOS long term memory**

**Fig. 2 Biological/CMOS long term memory**

Neurons are relatively efficient in terms of energy dissipation compared to historical CMOS because they judiciously regulate electricity. The result is an action potential with very low dissipative ($i^2R$) losses. Under the most idealistic of ideas there is reversible cycling of charges as well as of neurotransmitters, as in efficient rechargeable batteries. In terms of logic, a pulsating neuron may usually be replaced by a CMOS gate with dc logic levels. This is done so often that it is convenient to call it a *substitution principle*. Fig. 2b illustrates this principle for a long term memory circuit. Prior to activation the output is logic zero. Once triggered, this model holds forever.

Fig. 3a illustrates a biological short term neuron. It is convenient to think of a neuron as providing pulses whenever there is sufficient trigger potential within the dendrites where they attach to the soma. The triggering is prolonged because ions readily enter dendrites to produce a positive voltage but ions leave very slowly, thus enabling the dendrites to hold a triggering charge. Charge storage occurs because either 1) there is a natural shortage of internal potassium ions or 2) potassium ions are disabled by other chemicals possibly serotonin that converts into cyclic AMP (adenosine 5'-monophosphate). The end result either way is a prolonged burst of pulses to represent short term memory.

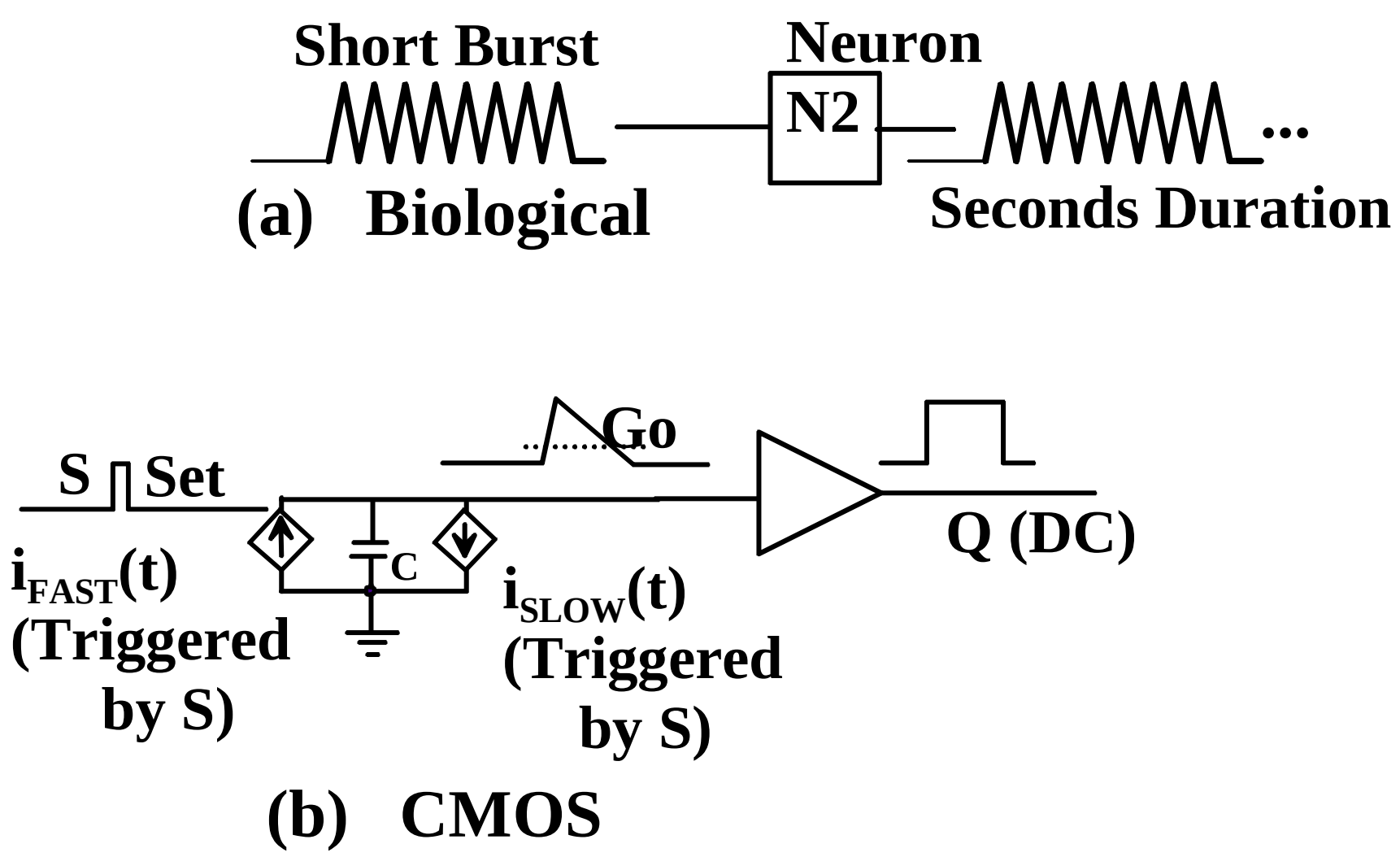

**Fig. 3 Biological/CMOS short term memory**

The CMOS analog for a short term memory neuron is shown in Fig. 3b. A pulse is generated by the difference between a fast current source $i_{FAST}(t)$ for a brief time (≈1 ms) and a slow current source $i_{SLOW}(t)$ for much longer (≈10 s). Charging and discharging *C* provides a *go signal* that a buffer converts into a single long pulse.

Short term memory neurons are conveniently modeled as having a one-to-one logical connection to respective attributes in Long term memory. Any designer can verify that it is possible to design an associative memory with search cues taken from any or all parts of the current image in short term memory. Simple circuits are available to permit a single recall even if there are multiple matches. Simple circuits also provide a memorization enable circuit requiring rehearsal in short term memory before writing into long term memory, just as in human memorization. What needs to be appreciated more is that associative memory enables an ongoing (artificially intelligent) machine as described below.

## Direction of attention

Mostly one's mind wanders from one topic to another chiefly in response to external stimuli coming in at random. Fig. 4 shows architecture involving *short* and *long term memory*. All sensory images are assumed to go into short term memory. But these images can be replaced quickly and are alternated with recalls from long term memory based on cues presented by the *cue editor*. The cue editor takes cues in short term memory and searches for matches in long term memory. Memory "hits" are assumed to have high importance, and might contain appropriate reactions, perhaps fight or flee in response to danger. New entries from long term memory soon dominate the lingering contents of short term memory.

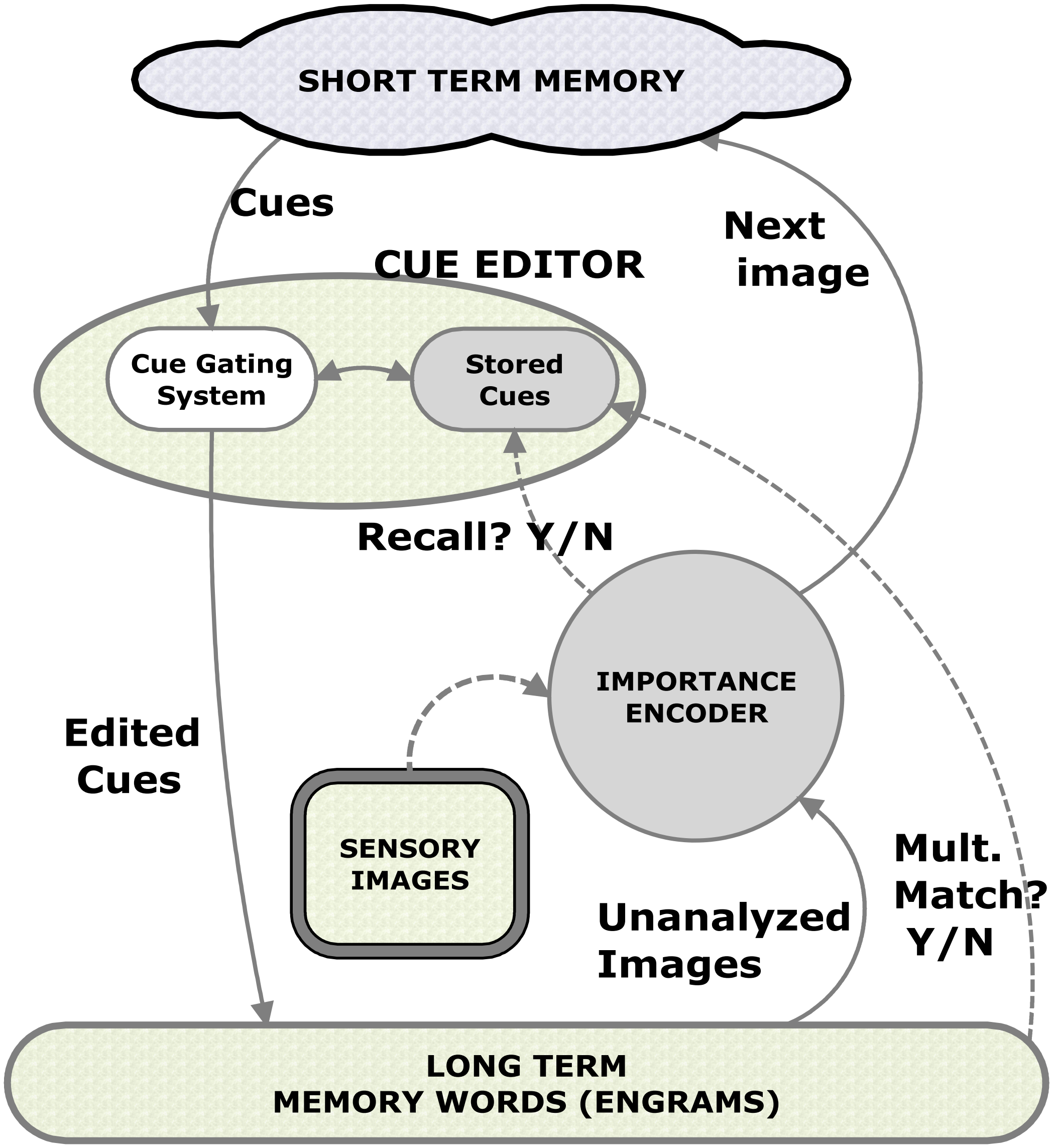

**Fig. 4 Artificial brain architecture**

**Simplified Editing of Inconsistent Cues** -- Sometimes an image in short term memory develops inconsistent cues, especially for imperfectly recalled specifics, so further recall on a given topic is impossible. During such a mental block there is a cue editing process that systematically removes cues and repeats the search for as many times as required, possibly at the maximum physical rate of many times per second. Once inconsistent cues are removed by the editor, a recall immediately occurs. Unexpectedly, the thing pops into consciousness. This often seems to be the case when trying to recall a name lost in long term memory.

Upon receiving a *no-recall* signal (Recall? Y/N in the figure) the cue editor springs into action, removes a cue and searches memory. If nothing returns, it replaces the removed cue…and then moves on, removing another. In hardware this is easily accomplished with a shift register counter, gating each cue as suggested in Fig. 5. This shows the unexpected simplicity of editing for recall as translated into CMOS logic gates. Theoretically two cues at a time could be removed, necessitating two shift registers, which takes longer; circuitry for this is not shown here. The underlying assumption for this hardware approach is that cues are limited in number.

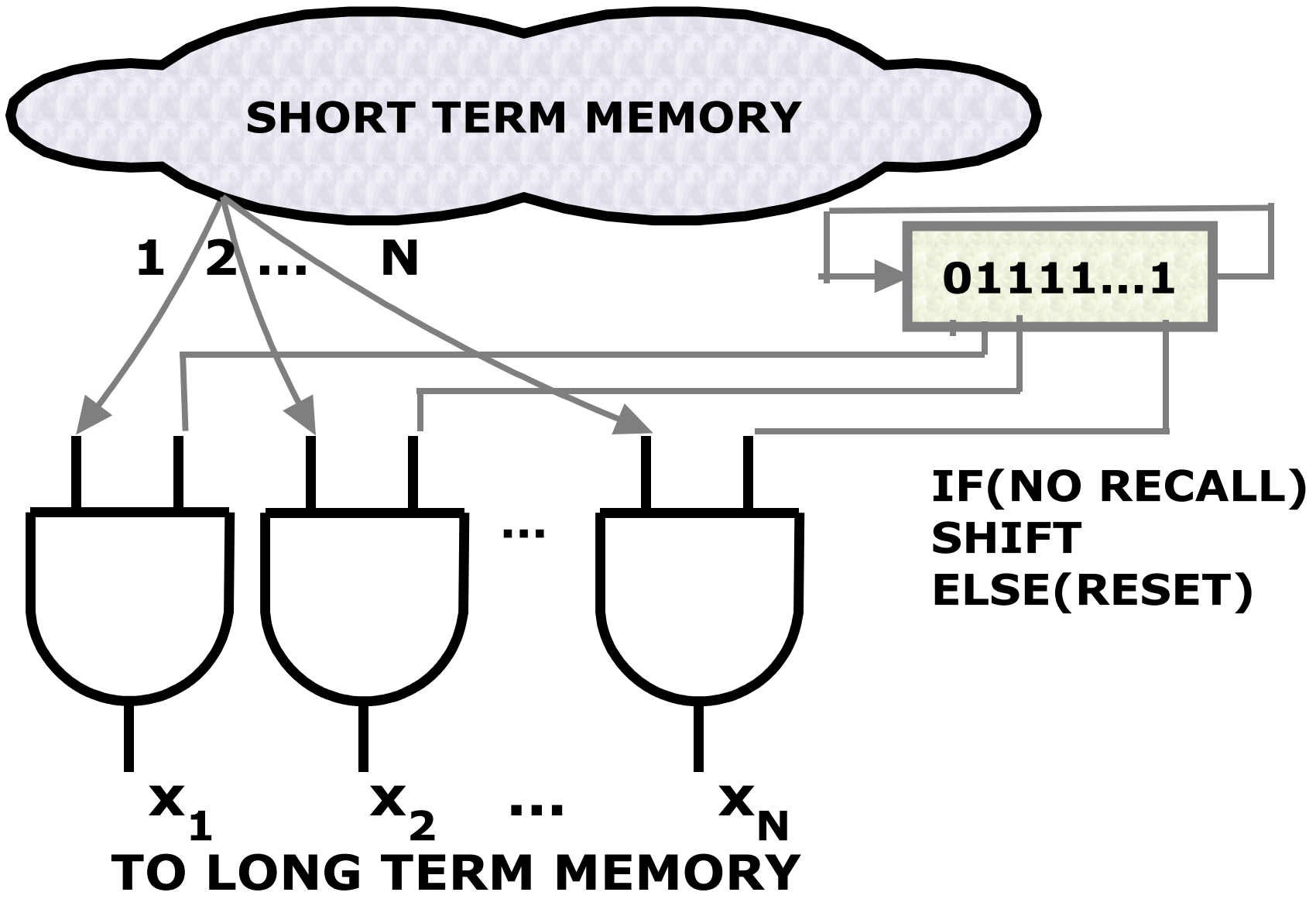

**Fig. 5 Cue editing simplified**

Simplified importance determinations – In the simplest possible model, the importance of individual recalls could be ignored completely. Just take anything that associates with the given cues and send it to consciousness.

On the other hand, characterizing recalls quantitatively is not difficult and may be accomplished in the *importance encoder*. Each recall has 1) a given count of cues used for memory search, 2) a given final count of attributes, and 3) a map of the locations of each attribute. Attributes can in principle be quantified with a combinational priority encoder, which is a standard digital circuit. For example, irrational attributes might be counted and, if they exceed a limit L, recall might be blocked, that is, repressed as in Sigmund Freud. Emotional illness and dreams seem to ignore L however. More elaborate calculations of importance (beyond excluding irrational memories) have been proposed but will not be bothered with here (Franklin 1995, Anderson 1983).

Note that systematic removal of cues is a way to *brainstorm*. During brainstorming a person experiences a wide variety of loosely related recalls, because fewer cues imply more matches in associative memory, and potentially creative recalls.

**Simplified Treatment of Multiple Matches** -- When there are too few cues, there can be multiple matches signaled subconsciously by a *multiple match line* (in the figure Mult. Match? Y/N). The simplest workable system is to permit into short term memory only the first match. Associative memory is easily designed to do this.

An interesting concept is to ignore importance and to pass each multiple match sequentially into short term memory. These would be experienced as rather confusing *differentiating* *multiple recalls* as in Richard Semon's memory nomenclature. Probably only the last in a stream of such recalls would be important enough to provide cues for future recalls. On the other hand, *non differentiating multiple recalls* would constitute an unambiguous recall, perhaps even a *resonance* or strong recall.

A more sophisticated approach would be to hold a given set of cues while as many matches as possible march through the importance encoder. Theoretically the encoder could gate into short term memory only that image with the highest importance. Eventually this process would be terminated when a new sensory image forces its way in.

## Mathematical thinking

Nearly everyone agrees that if a person can solve mathematical problems, that person is capable of thinking. Associative memory can be keyed with abstract information in equations as readily as with landmarks such as trees and stones. Consider a previously learned method identified as *method α* to be applied to solve for x, given visually on the blackboard: $2x + 5 = 11$. Abstractly, the equation is recognized to mean $A x + B = Y$. Its attributes are held in a single word of associative memory. Candidates for attributes are specific distributed locations for *2, x, +, 5, =,* and *11* and the form $A x + B = Y$ using method *α*.

Equations can be solved readily if a method to solve them has been learned. Learning is extremely important since it involves nearly everything humans do, not just mundane things like walking or memorizing phrases, but advanced things like mathematics. Artificial brains, as in models of biological brains, have learning in the form of an embedded state machine within long term memory (Burger, Sept 5, 2010). Fig. 6 depicts how the sight of this equation could activate a certain location in associative memory. This location is step 1 in a state machine sequence. A state machine, if uninterrupted, will step through its states without having to be directed by short term memory. In this sense it is efficient because each step does not have to be pondered consciously.

For the equation and method given, a state machine must accomplish the instructions listed. Not shown is code stored in long term memory to modify conscious short term memory. In general, an attribute (that is, a bit) has to move from one place to another in consciousness. A data bus is required, although it may be very simple. After operands have been moved into the arithmetic area there are two options: 1) look up and recover an arithmetic result from associative memory, or 2) compute an arithmetic result using nanocode supplied by the state machine. Option 1 is assumed in what follows.

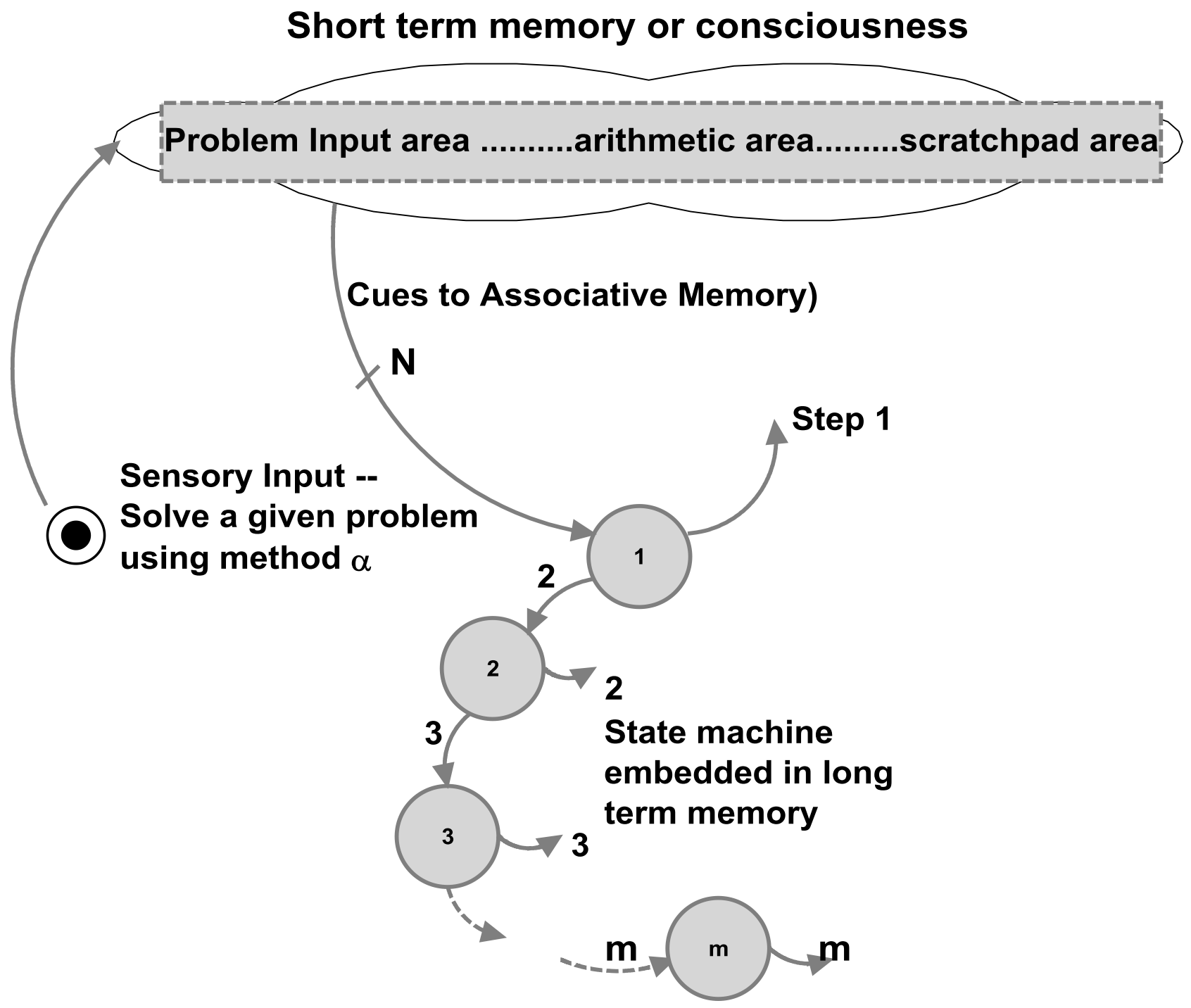

**Fig. 6 State machine for *2 x + 5 = 11***

**1. Move 11 to arithmetic area (leaving 0)**
**2. Move 5 to arithmetic area (leaving 0)**
**3. Look up associatively 11 - 5 = 6**
**4. Return result (6) to arithmetic area**
**5. Move 6 to the position of 11 in problem area**
**<u>2 x = 6 appears in problem area</u>**

Once *2 x = 6* is in short term memory, the first state machine may end. An associative search now locates a procedure as in Fig. 7 to solve *2 x = 6*. Mathematics obviously may increase in complexity, meaning more states. Fig. 8 illustrates programming to solve a simple differential equation.

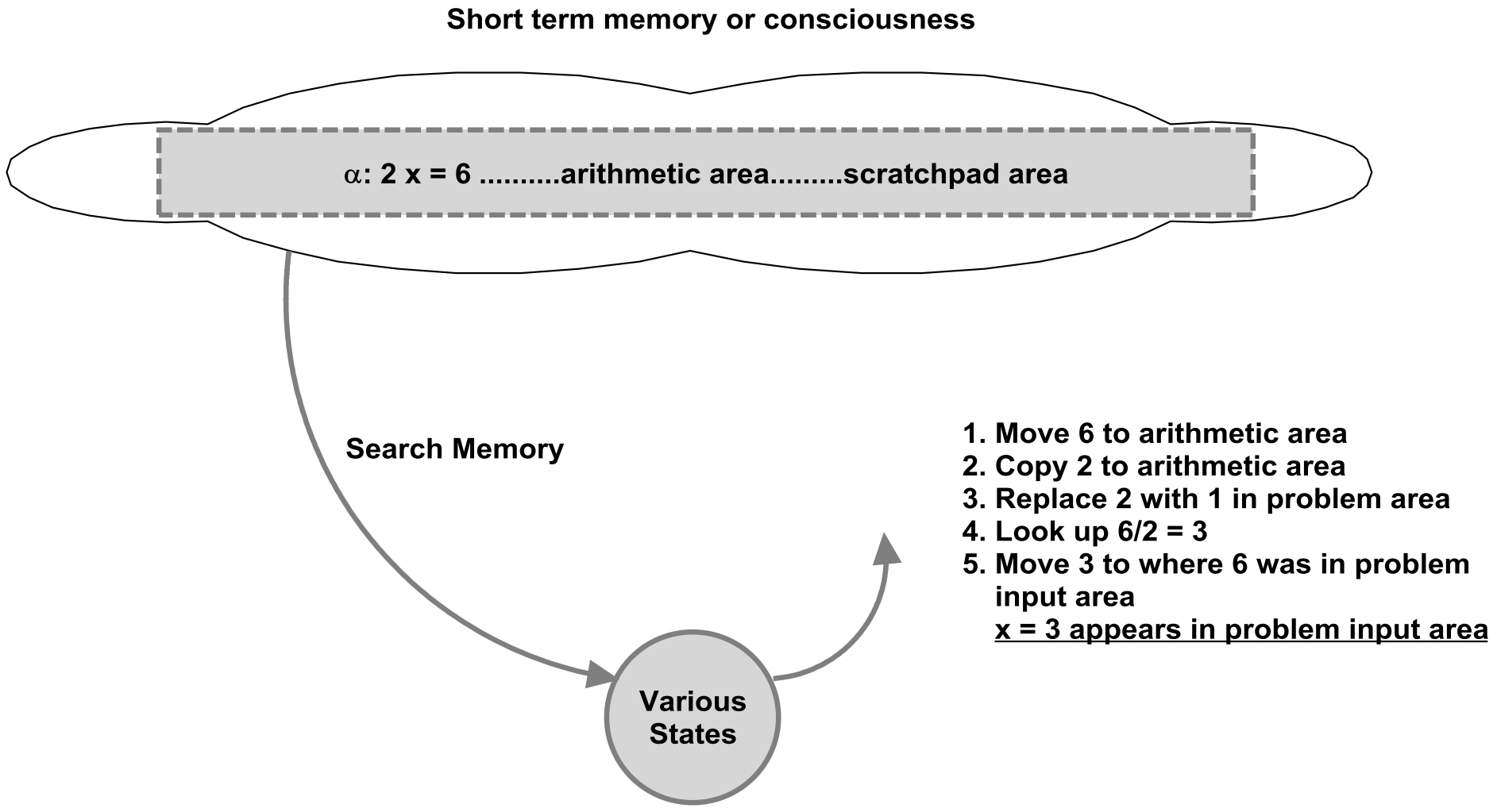

**Fig. 7 Solving *2 x = 6***

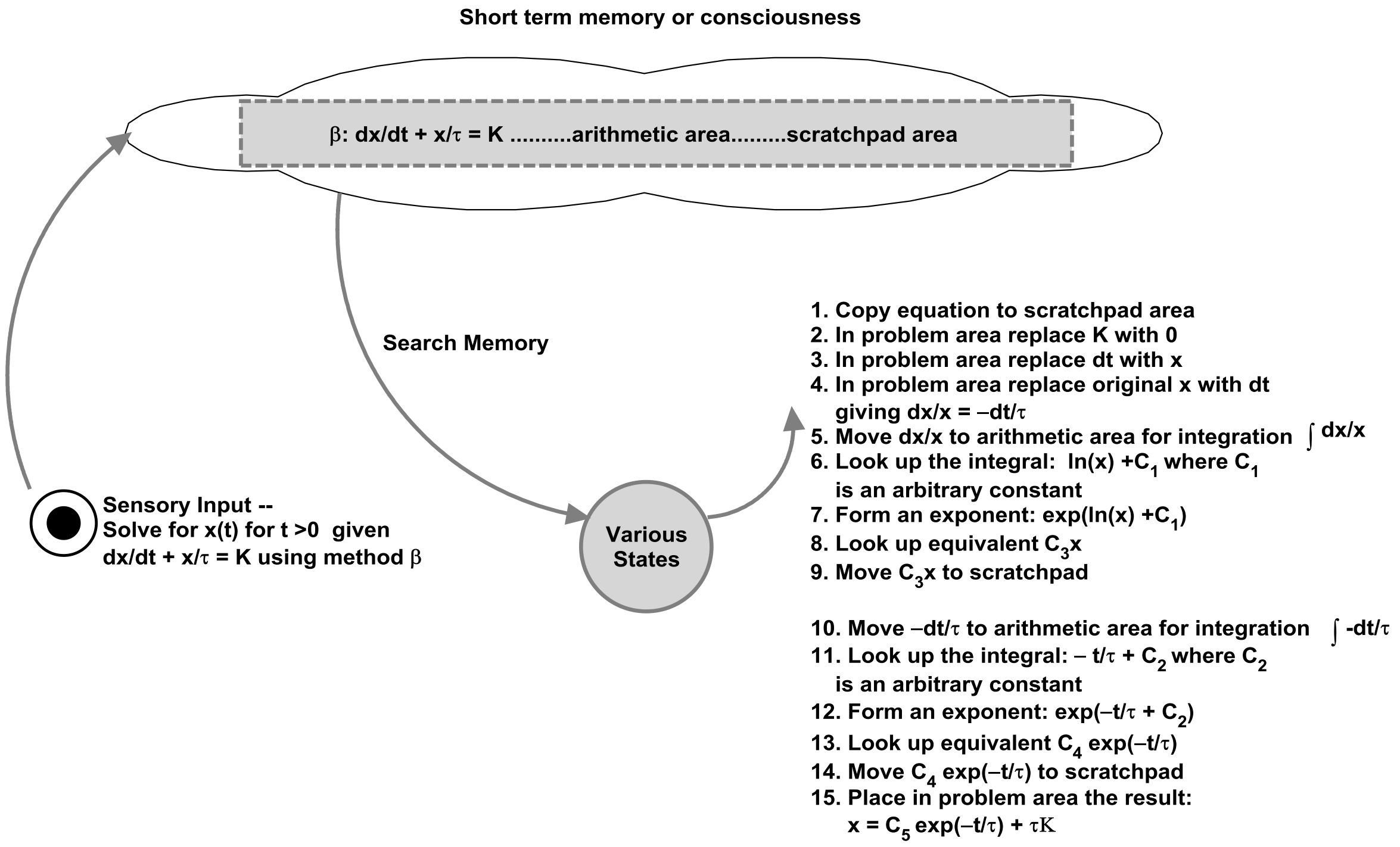

**Fig. 8 Thinking through a differential equation** ***dx/dt*** **+** ***x/τ*** **=** ***K***

# Finding One's Way

Finding one's way requires a sort of thinking. Landmarks are signs in the physical world. For instance, assume that a hiker enters into a large black forest with unmarked trails. At each intersection choices are necessary. At the first intersection, for example, are 1 clump of bushes, two conifer trees, 1 deciduous tree and 1 stone. He or she turns right, which might be coded as 1 B, 2 C, 1 D, 1 S / *Right*, where B = Bushes; C = Conifers; D = deciduous; S = Stones. With effort, landmarks may be remembered. After doing the same for each intersection the hiker arrives at a black forest beer stand. A mathematical record, assuming four intersections might be expressed as follows:

$$\begin{bmatrix} 1 & 2 & 1 & 1 \\ 0 & 2 & 2 & 2 \\ 0 & 2 & 1 & 2 \\ 1 & 2 & 1 & 2 \end{bmatrix} \begin{pmatrix} B \\ C \\ D \\ S \end{pmatrix} \quad \begin{matrix} /\,Right \\ /\,Left \\ /\,Straight \\ /\,Left \end{matrix}$$

Each word (row) of memory includes the direction taken. The determinant of the matrix is non zero, indicating that each intersection has unique markers. To go home, a look at Fig. 9 reveals visual cues as follows: 1 B, 2 C, 1 D and 2 S. This exactly matches row 4 of the matrix and enables a recall. But what if the hiker drank too much and does not see all landmarks? For example, an interesting situation occurs if the hiker can no longer see bushes. Then cues become 2C, 1D, 2S and memory is underdetermined. In this case of *Mult. Match*, rows 3 and 4 of the matrix with 0B and 1B come to mind. Without additional information one could very well take a wrong path and vanish in the wilderness.

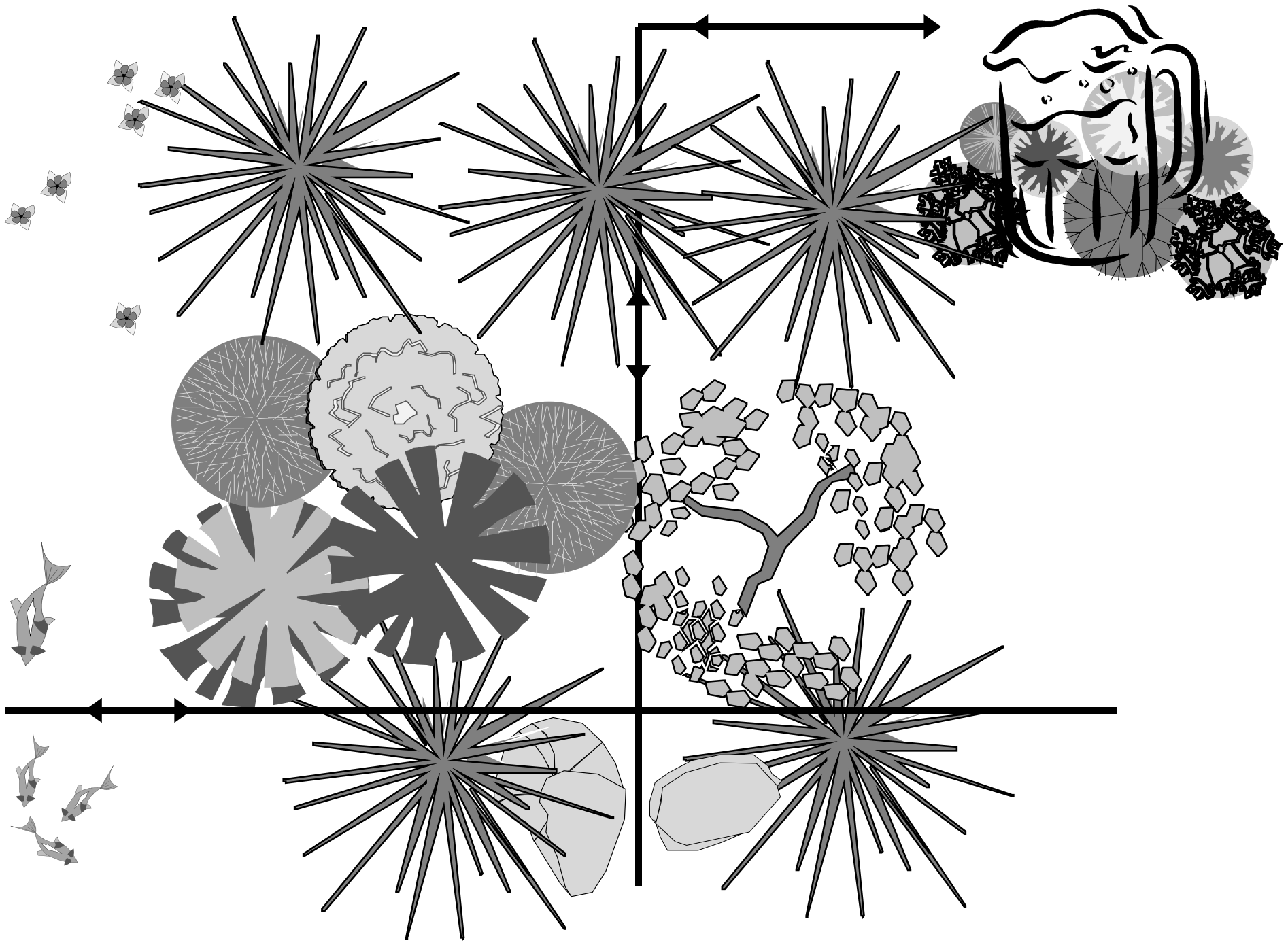

**Fig. 9 End of the trail in a black forest**

Another interesting case occurs if our inebriated hiker has a mental block caused by an imagined but incorrect cue. For example, the hiker may notice little fishes on his way home that he did not see on the way in. When cues are inconsistent an editing process is evoked, because a *No Recall* signal occurs subconsciously. Over a period of time cues are removed (subliminally, one at a time) to help locate the desired memory. Once little fishes are removed, recall occurs immediately.

But to avoid getting lost, a hiker must do more than memorize. When back-tracking to return home, left must be converted to right and vice versa. As any dance teacher knows well, right and left are not just pondered words; actions relating to left and right are understood prior to their mental verbalization. Once left and right are properly learned, an embedded state machine can subliminally interchange left and right in consciousness. The process is illustrated in Fig. 10.

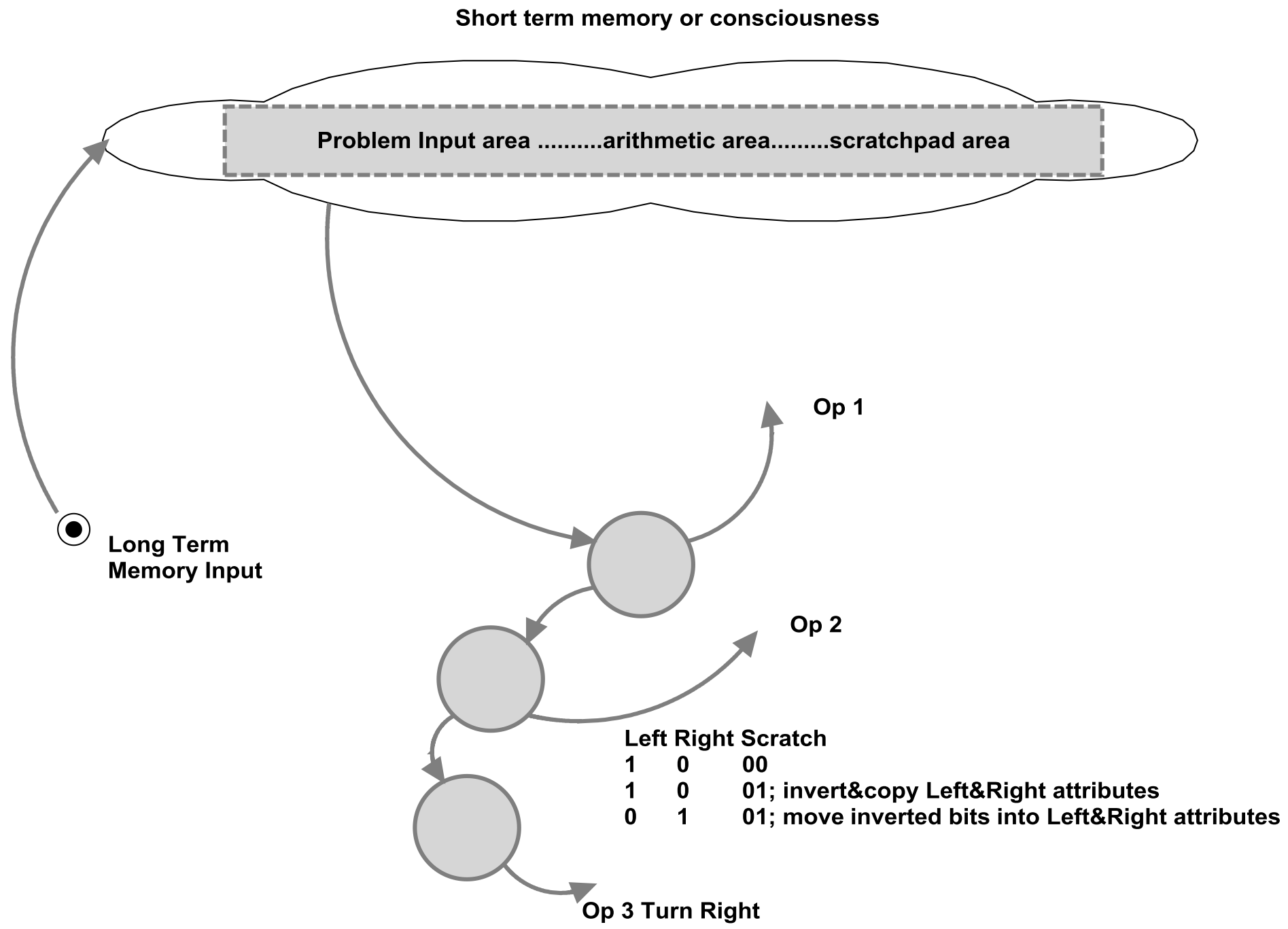

**Fig. 10 Interchanging Left and Right**

# Artificial Nanoprocessing

Everyone has rudimentary processing capability within the mathematical sections of their short term memories because they need to process attributes. This is easily accomplished artificially using multiplexing in a bus system to move bits. One type of implementation uses a single wire to link nanoprocerssors (Burger 2009). This type of implementation assumes electrically reversible CMOS and a superior type of programming known as reversible programming, because it minimizes heat release. Jaded designers are of course welcome to use what is more popular today, fast, hot-running serial logic with dissipative one-way programming.

Electrically reversible CMOS and logically reversible programming are going to be used in the example below. This example is being promoted since any fabrication approaching the human neural density must be efficient since otherwise heat release would be excessive. Reversible processing for short term memory can be approximated using CMOS transistors in a system called SAML, which here means Self-analyzing memory logic (Burger 2007, 2009). Under this reversible system, one or more bits are selected by a TO signal. These bits may be flipped by the truth value of other bits in short term memory. These controlling bits are identified by a FM signal. If they are all true, TO bits are flipped. It turns out that TO and FM bits have to be disjoint for reversible programming. Fig. 11 diagrams a plan to reduce $2x + 5 = 11$ assuming a byte (4 bits) for each attribute. Note the binary code. Biological memory, of course, uses a different code, but for a CMOS realization, binary is convenient.

Note that a single bus connects all memory elements within the mathematical section of short term memory. The TO and FM signals are applied from a state machine within long term memory. Fig. 12 illustrates the words (or engrams) of nanocode in a state machine.

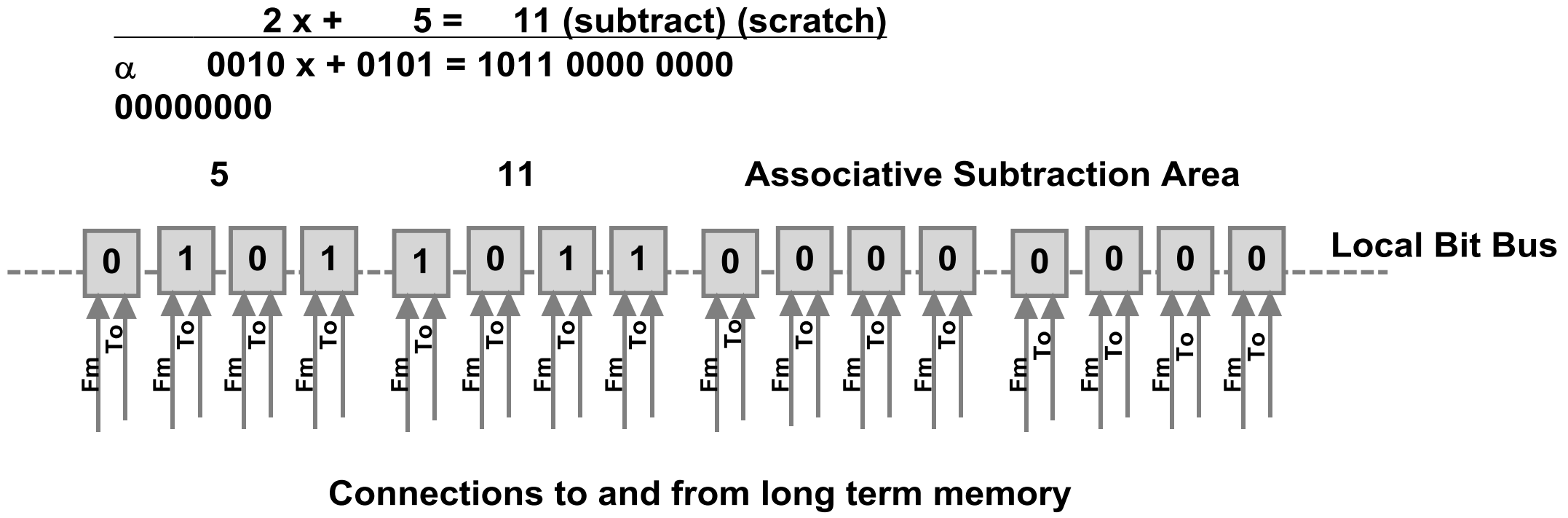

**Fig. 11 Nanoprocessors applied to reduce 2 x + 5 = 11 to 2 x = 6**

To begin, the LSB, that is, the rightmost bit under 1011 becomes a FM signal. The LSB under the first byte of the *subtract area* becomes a TO signal. In this way the first bit of the eleven moves into the subtract area. After three more FM-TO operations the eleven is copied into the left four bits of the associative subtraction area. The next four operations replace the original eleven with zero, completing moving eleven to the field of the first operand. This is accomplished by the first 8 operations, or the first 16 lines in Fig. 12.

Continuing the fun, the five is now moved to be the second operand in the subtract area. This is accomplished by the next 16 lines in Fig. 12. At this point it is assumed that the difference between 11 and 5 is found within associative memory. Note that the difference could have been calculated within short term memory using SAML, but it is unnecessary to do this when associative memory directly returns a difference. The next four operations (next 8 lines in Fig. 12) copy the difference back to where the eleven originally was in the problem input area.

# Conclusions

Sensory signals and associated remembrances trigger a subconscious process to generate new thoughts. This circular causal relationship is necessary and sufficient for a cybernetic perspective. Orchestrating the interplay between conscious short term and subconscious associative memory is a proposed system of subliminal processing. It seems as though one cannot trace one's own mental processes given that they are subliminal. Even so, little pieces of the brain puzzle have fitted together over the years so that now substantial progress is possible via circuit synthesis.

```
          2 x +      5 =     11 (subtract) (scratch)
α      0010 x + 0101 = 1011 0000 0000 00000000; coded
result

Fm     0000 0 0 0000 0 0001 0000 0000 00000000; move 11
To     0000 0 0 0000 0 0000 0001 0000 00000000
Fm     0000 0 0 0000 0 0010 0000 0000 00000000
To     0000 0 0 0000 0 0000 0010 0000 00000000
Fm     0000 0 0 0000 0 0100 0000 0000 00000000
To     0000 0 0 0000 0 0000 0100 0000 00000000
Fm     0000 0 0 0000 0 1000 0000 0000 00000000
To     0000 0 0 0000 0 0000 1000 0000 00000000

Fm     0000 0 0 0000 0 0000 0001 0000 00000000; repl 11 by 0
To     0000 0 0 0000 0 0001 0000 0000 00000000
Fm     0000 0 0 0000 0 0000 0010 0000 00000000
To     0000 0 0 0000 0 0010 0000 0000 00000000
Fm     0000 0 0 0000 0 0000 0100 0000 00000000
To     0000 0 0 0000 0 0100 0000 0000 00000000
Fm     0000 0 0 0000 0 0000 1000 0000 00000000
To     0000 0 0 0000 0 1000 0000 0000 00000000

Fm     0000 0 0 0001 0 0000 0000 0000 00000000; move 5
To     0000 0 0 0000 0 0000 0000 0001 00000000
Fm     0000 0 0 0010 0 0000 0000 0000 00000000
To     0000 0 0 0000 0 0000 0000 0010 00000000
Fm     0000 0 0 0100 0 0000 0000 0000 00000000
To     0000 0 0 0000 0 0000 0000 0100 00000000
Fm     0000 0 0 1000 0 0000 0000 0000 00000000
To     0000 0 0 0000 0 0000 0000 1000 00000000

Fm     0000 0 0 0000 0 0000 0000 0001 00000000; repl 5 by 0
To     0000 0 0 0001 0 0000 0000 0000 00000000
Fm     0000 0 0 0000 0 0000 0000 0010 00000000
To     0000 0 0 0010 0 0000 0000 0000 00000000
Fm     0000 0 0 0000 0 0000 0000 0100 00000000
To     0000 0 0 0100 0 0000 0000 0000 00000000
Fm     0000 0 0 0000 0 0000 0000 1000 00000000
To     0000 0 0 1000 0 0000 0000 0000 00000000

α      0010 x + 0000 = 0000 1011 0101 00000000; coded
result
α      0010 x + 0000 = 0000 0000 0110 00000000; recall dif

Fm     0000 0 0 0000 0 0000 0000 0001 00000000; move dif
To     0000 0 0 0000 0 0001 0000 0000 00000000
Fm     0000 0 0 0000 0 0000 0000 0010 00000000
To     0000 0 0 0000 0 0010 0000 0000 00000000
Fm     0000 0 0 0000 0 0000 0000 0100 00000000
To     0000 0 0 0000 0 0100 0000 0000 00000000
Fm     0000 0 0 0000 0 0000 0000 1000 00000000
To     0000 0 0 0000 0 1000 0000 0000 00000000

α      0010 x + 0000 = 0110 0000 0110 00000000; coded
result
```

COMMANDS

**Fig. 12 Nanocode to arrive at 2 x = 6 in short term memory**

It was determined early in this project that artificial neural networks are hopelessly slow and generally inappropriate for memory operations involving recalling, thinking and learning. What exists biologically are Boolean neurons, including short term memory neurons and long term memory neurons (a biological neuron does not necessarily have myriad weighted inputs, a linear summation node and a nonlinear comparator sometimes mistakenly termed a *neural net*). To move closer to biological reality with a glimmer of hope for designing an anthropomorphic brain, a substitution principle was developed in which dc logic can be used to replace the pulsating logic of biological neurons. Under this principle, most biological neurons map homomorphically into CMOS gates.

Architecture is provided above in which state machines embedded deep within associative long term memory are permitted to execute previously learned procedures and actions. Beyond everyday walking and reciting and so forth, mathematical equations can be solved, which is a form of thinking. To solve equations, embedded state machines systematically apply coded instructions to short term memory to modify attributes and to give the impression of thinking.

In conclusion, this paper attempts to contribute to an authentic frontier of knowledge: The circuitry for a human brain. This work is necessarily humble because it is an example of circuit synthesis not for money but rather for common knowledge. It is sincerely hoped that someone somewhere will be enlightened and appreciative in some small way for the ideas presented above.

## Citations